\documentclass[sigconf]{acmart}
\usepackage{graphicx}
\usepackage{amsthm}
\usepackage{multirow}

\usepackage{xcolor}
\usepackage{xspace}
\usepackage[ruled,vlined,linesnumbered]{algorithm2e}
\usepackage{booktabs,tabularx}
\usepackage{graphicx}  % 插图宏包
\usepackage{subcaption}
\usepackage{caption}   % 可选：更好地控制标题样式
\usepackage{float}     % 可选：允许使用 [H] 强制位置
\usepackage{enumitem}
\setlist{nosep} % 等价于 itemsep=0pt, topsep=0pt, parsep=0pt, partopsep=0pt

\newcommand{\eg}{\emph{e.g.},\xspace}
\newcommand{\ie}{\emph{i.e.},\xspace}

\usepackage[most]{tcolorbox}
\newcolumntype{Y}{>{\raggedright\arraybackslash}X}
\tcbset{
  profilebase/.style={
    enhanced, breakable,
    boxrule=0.6pt, arc=2mm,
    left=2mm,right=2mm,top=1.5mm,bottom=1.5mm,
    before skip=6pt, after skip=8pt,
    fonttitle=\bfseries, coltitle=white,
    boxed title style={sharp corners, boxrule=0pt},
    attach boxed title to top left={xshift=2mm,yshift*=-1.2mm},
  }
}

\newtcolorbox{profilebox}[2][]{profilebase, title={#2}, #1}
\definecolor{ForestGreen}{RGB}{34,139,34}  % #228B22
\definecolor{RoyalBlue}{RGB}{65,105,225}   % #4169E1
\definecolor{DarkGray}{RGB}{169,169,169}   % #A9A9A9
\definecolor{DimGray}{RGB}{105,105,105}    % #696969

\usepackage{siunitx}
\AtBeginDocument{%
  }

\setcopyright{acmlicensed}
\copyrightyear{2018}
\acmYear{2018}
\acmDOI{XXXXXXX.XXXXXXX}

\acmConference[Conference acronym 'XX]{Make sure to enter the correct
  conference title from your rights confirmation email}{June 03--05,
  2018}{Woodstock, NY}

\acmISBN{978-1-4503-XXXX-X/2018/06}

\begin{document}

\title{AgentBalance: Backbone-then-Topology Design for Cost-Effective Multi-Agent Systems under Budget Constraints}

\author{Shuowei Cai}
\email{scaiak@connect.hkust-gz.edu.cn}
\affiliation{%
  \institution{The Hong Kong University of Science and Technology
(Guangzhou)}
 \country{}
}

\author{Yansong Ning}
\email{yning092connect.hkust-gz.edu.cn}
\affiliation{%
  \institution{The Hong Kong University of Science and Technology
(Guangzhou)}
 \country{}
}

\author{Hao Liu}
\email{liuh@ust.hk}
\affiliation{%
  \institution{The Hong Kong University of Science and Technology
(Guangzhou)}
 \country{}
}

% \affiliation{%

% \affiliation{%

% \affiliation{%

% \affiliation{%

% \affiliation{%

% \affiliation{%

% \affiliation{%

% \affiliation{%

\renewcommand{\shortauthors}{Trovato et al.}

\begin{abstract}
Large Language Model (LLM)–based multi-agent systems (MAS) have become indispensable building blocks for web-scale applications (\eg web search, social network analytics, online customer support), with cost-effectiveness becoming the primary constraint on large-scale deployment.
While recent advances seek to improve MAS cost-effectiveness by shaping inter-agent communication topology and selecting agent backbones, they seldom model and optimize under explicit token-cost and latency budgets that reflect deployment constraints, leading to topology-first designs and suboptimal cost-effectiveness under budget constraints.
In this paper, we present \textbf{\textsc{AgentBalance}}, a framework for constructing cost-effective MAS under explicit token-cost and latency budgets via a backbone-then-topology design.
Specifically, we first propose a \emph{backbone-oriented agent generation} module that constructs agents with heterogeneous backbones via LLM pool construction, pool selection, and agent role-backbone matching.
Then, we propose an \emph{adaptive MAS topology generation} module that guides inter-agent communication through agent-representation learning, gating, and latency-aware topology synthesis.
Extensive experiments on benchmarks with 14 candidate LLM backbones show that \textsc{AgentBalance} delivers up to 10\% and 22\% performance gains under matched token-cost and latency budgets, respectively, and achieves strong AUCs across benchmarks on performance–budget curves. It also works as a plug-in for existing MAS, further improving performance under the same token-cost and latency constraints, and exhibits strong inductive ability on unseen LLMs for practical, budget-aware deployment. Our code can be found at \url{https://github.com/usail-hkust/AgentBalance_}
\end{abstract}

\begin{CCSXML}
<ccs2012>
 <concept>
  <concept_id>00000000.0000000.0000000</concept_id>
  <concept_desc>Do Not Use This Code, Generate the Correct Terms for Your Paper</concept_desc>
  <concept_significance>500</concept_significance>
 </concept>
 <concept>
  <concept_id>00000000.00000000.00000000</concept_id>
  <concept_desc>Do Not Use This Code, Generate the Correct Terms for Your Paper</concept_desc>
  <concept_significance>300</concept_significance>
 </concept>
 <concept>
  <concept_id>00000000.00000000.00000000</concept_id>
  <concept_desc>Do Not Use This Code, Generate the Correct Terms for Your Paper</concept_desc>
  <concept_significance>100</concept_significance>
 </concept>
 <concept>
  <concept_id>00000000.00000000.00000000</concept_id>
  <concept_desc>Do Not Use This Code, Generate the Correct Terms for Your Paper</concept_desc>
  <concept_significance>100</concept_significance>
 </concept>
</ccs2012>
\end{CCSXML}


\ccsdesc[500]{Computing methodologies~Multi-agent systems}
\ccsdesc[300]{Computing methodologies~Natural language processing}
\ccsdesc[300]{Computing methodologies~Machine learning}
% \ccsdesc[100]{Information systems~World Wide Web}

% \ccsdesc[300]{Information systems~World Wide Web~Web applications}
% \ccsdesc[100]{Software and its engineering~Software performance}

%\keywords{Multi-agent System, Cost-effectiveness.}
\keywords{Multi-agent System, Heterogeneous Large Language Models, Cost-effective, Budget Constraint.}

\received{20 February 2007}
\received[revised]{12 March 2009}
\received[accepted]{5 June 2009}

\maketitle
\begin{figure}%[h] % h=here, t=top, b=bottom, p=page, H=exactly here
    \vspace{0.4cm}
    \centering
    \includegraphics[width=0.4\textwidth]{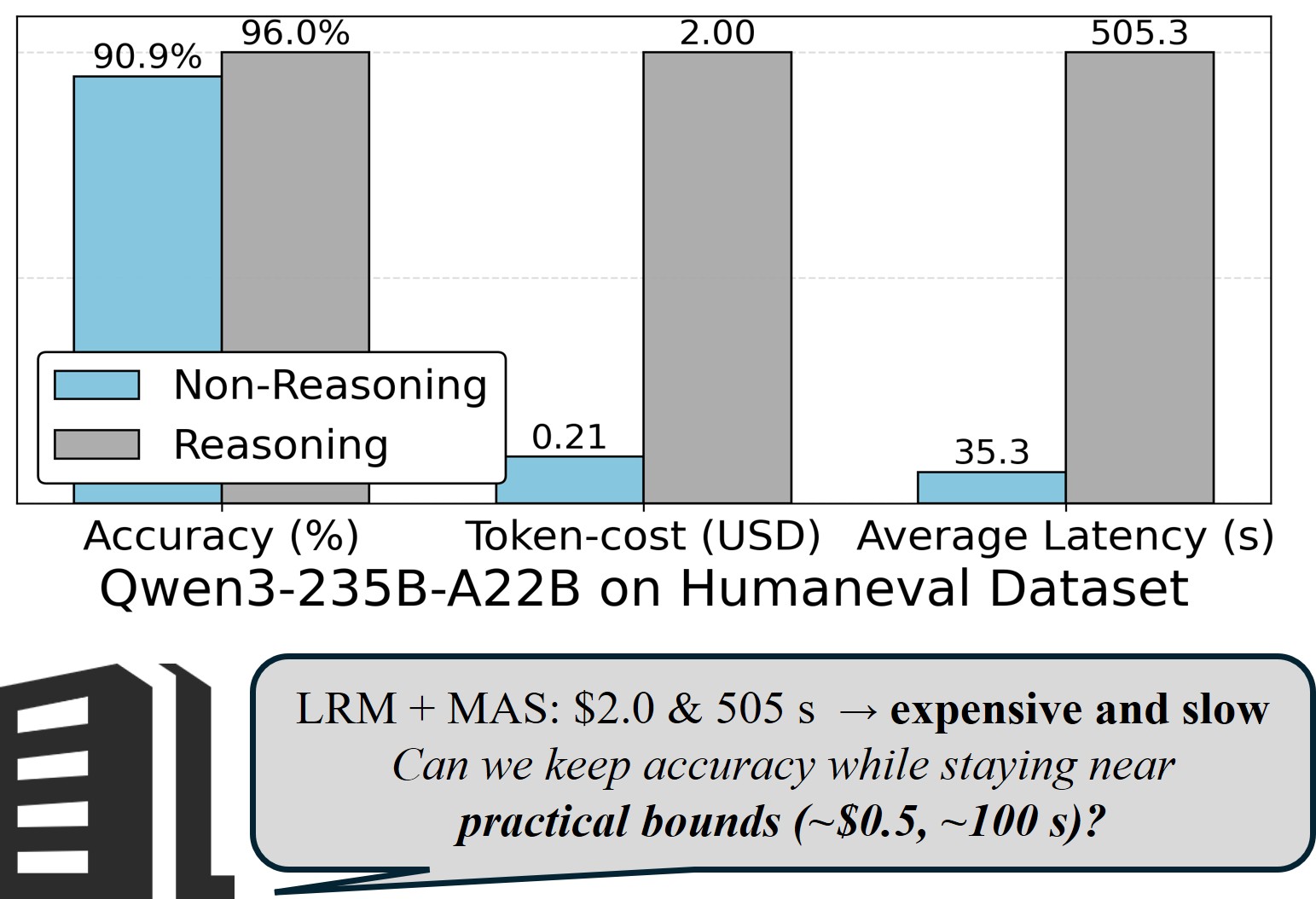}  
    \vspace{-0.3cm}
\caption{High performance in MAS often comes with token-cost and latency that exceed practical budgets. Stakeholders seek solutions that achieve competitive accuracy while respecting token-cost and latency budgets, motivating budget-aware, cost-effective MAS.}
    \vspace{-0.4cm}
\label{fig:intro1}
\end{figure}

\section{Introduction}
\begin{figure*}%[h] % h=here, t=top, b=bottom, p=page, H=exactly here
    \centering
    \includegraphics[width=0.9\textwidth]{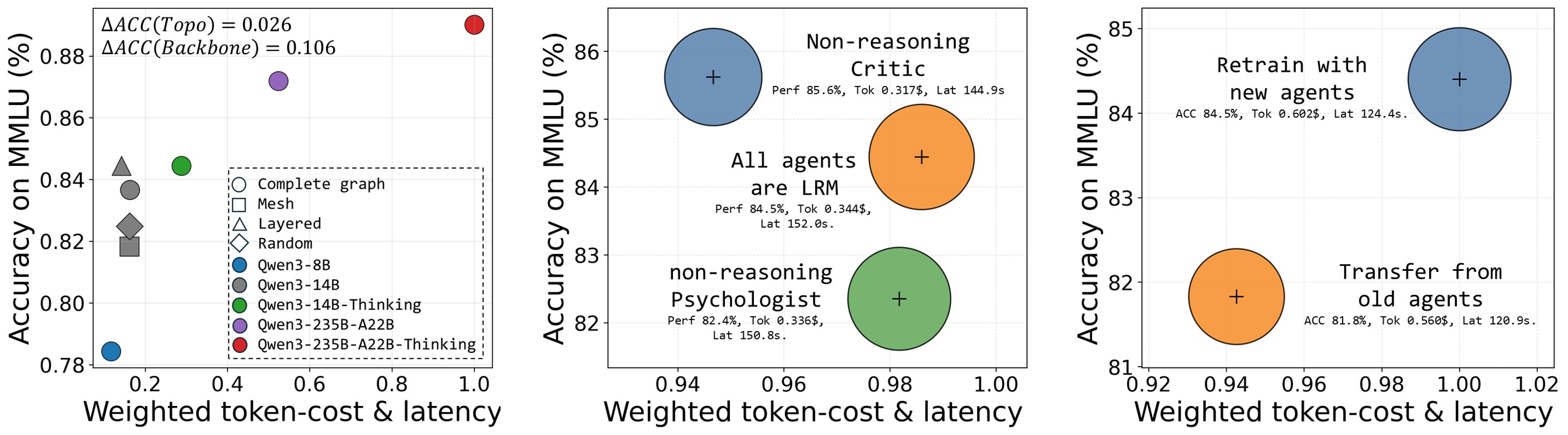}
    \vspace{-0.3cm}
    \caption{
\textbf{(Left)} Backbone choice moves the cost–performance frontier more than topology modifications in MAS.
\textbf{(Middle)} With topology fixed, assigning a non-reasoning model (vs.\ an LRM) to a suitable role (\eg \emph{Critic}) improves cost-effectiveness, whereas role–backbone mismatches (\eg \emph{Psychologist}) degrade it.
\textbf{(Right)} The optimal communication topology is backbone-dependent; after changing agent backbones, re-synthesizing the topology for the new agents outperforms transferring the old one.
}
    \label{fig:intro}
    \vspace{-0.4cm}
\end{figure*}

LLM-based multi-agent systems (MAS) have demonstrated strong performance in web-scale applications such as web search~\citep{chen2025ma4div}, social-network forecasting~\citep{miyake2024netevolved}, and online analytics~\citep{rezaei2025pipemind}, by decomposing complex tasks into specialized roles~\citep{metagpt}, integrating tool use and web APIs~\citep{toolformer}, and enabling inter-agent collaboration~\citep{autogen,camel}.
However, such coordination structurally couples system performance to token-cost (via API token consumption and inference-time computation~\citep{token_LLM}) and end-to-end latency: deeper interaction chains trigger more LLM calls, lengthen inter-agent contexts, and add serialization overhead~\citep{token_LLM}.
Equipping agents with advanced Large Reasoning Models (LRMs), \eg OpenAI o3 and DeepSeek\mbox{-}R1~\citep{openai-o3,deepseek-r1}, as backbones further amplifies this coupling in MAS~\citep{LRM-MAS1,LRM-MAS2,X-MAS}.
As shown in Figure~\ref{fig:intro1}, while accuracy improves, the reasoning-heavy decoding and longer outputs of LRMs increase completion length and inference time~\citep{longshort,LRM_latency1}, driving token-cost and latency to levels that are hard to sustain in practice.
These elevated costs render many configurations impractical for production web services, \eg ride-hailing dispatch~\citep{dima}, contact centers~\citep{LLM_Contact_Center}, and real-time social agents~\citep{lyfe_agents}, where token-cost and latency are governed by explicit budgets or operational constraints.
Accordingly, optimizing for performance alone is misaligned with deployment realities. \emph{MAS should be designed and evaluated in a budget-aware manner, maximizing performance subject to token-cost and latency constraints to improve cost-effectiveness.}

Despite growing interest in powerful MAS~\citep{multiagentdebate,offlineLLM,LLM-blender,gptswarm,hugginggpt}, existing studies inadequately address budget-constrained settings on both the objective and methodology fronts.
On the objective side, many works prioritize performance, treating token-cost as secondary and rarely emphasizing latency. They also seldom evaluate under explicit token-cost and latency budgets, leaving the practical deployability of these methods under realistic constraints unclear.
On the methodology side, prior work concentrates on inter-agent communication~\citep{AgentPrune,G-Designer,agentdropout}, reducing communication redundancy or removing redundant agents for cost-effectiveness while typically assuming a single strong backbone, thereby overlooking how backbone size, family, and type shape cost-effectiveness.
Very recent work such as \textsc{MasRouter} incorporates multiple LLM backbones for MAS~\citep{masrouter}, but remains topology-first and does not account for two properties of backbone choice for agents: (i) backbone choice is a primary driver of movement along the frontier between cost and performance relative to topology modification (Figure~\ref{fig:intro}, left), and (ii) backbone choice shapes the optimal communication topology (Figure~\ref{fig:intro}, right).
Taken together, these observations motivate a backbone-first strategy: fix the backbone to define the feasible performance region under given budgets, then optimize topology within it. We therefore adopt a backbone-then-topology approach to construct cost-effective MAS under budget constraints. However, this design is nontrivial and leads to two complementary challenges:

\textbf{(1) Constructing cost-effective agents with heterogeneous backbones.}
Assigning backbones to agents given the context can improve cost-effectiveness (\eg a mid-sized non-reasoning LLM often suffices for simple tasks such as tool invocation~\citep{frugalgpt}).
However, this is nontrivial in MAS: the set of candidate backbones is large and heterogeneous (sizes from billions to hundreds of billions; reasoning and non-reasoning), spanning a wide frontier between cost and performance.
Appropriate assignment depends jointly on the query and the agent role, with mismatched assignments markedly degrading performance (Figure~\ref{fig:intro}, middle).
Thus, how to assign backbones precisely so that each agent’s capability matches task demands within budget constraints is the first challenge. 

\textbf{(2) Designing communication topology for agents with heterogeneous backbones.}
Designing role-aware communication patterns yields joint gains in performance and token-cost~\citep{AGP-arxiv,MAS_topo,EIB-learner}.
However, constructing a latency-aware communication topology for agents with heterogeneous backbones is nontrivial: the optimal topology of a MAS depends on the agents’ backbones (Figure~\ref{fig:intro}, right).
In addition, backbone heterogeneity across roles and queries exacerbates this issue, making it substantially harder to estimate each agent’s impact in the system and to decide with whom to communicate under latency considerations.
Therefore, how to estimate each agent’s marginal contribution and adaptively design a latency-aware communication topology within budget constraints is the second challenge.
% \textbf{(2) How to design latency-aware communication topology on agents with heterogeneous backbones.}
% Designing role-aware communication patterns yield joint gains in performance and token-cost~\citep{AGP-arxiv,MAS_topo,EIB-learner}, but constructing a latency-aware topology for agents with heterogeneous backbones remains unexplored. The optimal communication structure is backbone-dependent (Figure~\ref{fig:intro}, Right), and backbone heterogeneity across roles and queries markedly increases the difficulty of topology construction.
% Therefore, the second challenge is to estimate each agent’s query- and backbone-conditioned marginal contribution and to \emph{adaptively} schedule when and with whom it communicates, synthesizing a topology that reduces latency as well as token-cost.

To address the aforementioned challenges, we propose \textsc{\textbf{AgentBalance}}, a unified framework for constructing cost-effective MAS with a backbone-then-topology design under token-cost and latency budgets.
Specifically, we first propose a \emph{Backbone-Oriented Agent Generation} module for heterogeneous agent generation.
By sequentially performing LLM pool construction with profiling, difficulty-aware pool selection and query-conditioned role-backbone matching, we allocate an appropriate backbone for each agent based on the corresponding token-cost and latency budgets.
Moreover, we propose an \emph{Adaptive MAS Topology Generation} module to guide the communication between agents with heterogeneous backbone. Through unified agent representation learning, agent gating and latency-aware communication topology synthesis, we construct a cost-effective MAS.
Finally, by end-to-end optimization under token-cost and latency penalties, we achieve joint balance across performance, token-cost, and latency.
Our contributions are:
\begin{enumerate}[leftmargin=*, itemsep=0.25ex, topsep=0.2ex, parsep=0pt]
\item \textbf{Problem Formulation:}
We formalize cost-effectiveness in MAS as a tri-objective view of performance, token-cost and end-to-end latency, highlighting evaluation with explicit token-cost and latency budget requirements beyond performance alone for large-scale web-based applications.
\item \textbf{Methodology:}
We present \textsc{AgentBalance}, a framework for constructing cost-effective MAS with a backbone-then-topology design. By first constructing agents via backbone-oriented agent generation and then composing the whole system with adaptive MAS topology generation, \textsc{AgentBalance} achieves strong performance under explicit token-cost and latency budgets.
\item \textbf{Experimental Validation:} We instantiate MAS using \textsc{AgentBalance} with 14 heterogeneous LLM backbones and evaluate it against four baselines across benchmarks in three domains. Under matched token-cost and latency budgets, \textsc{AgentBalance} achieves performance gains of up to 10\% and 22\%, respectively. It also serves as a plug-in to enhance existing MAS and exhibits strong inductive ability in adapting to unseen LLMs.
\end{enumerate}

\section{Background}
\paragraph{Multi-agent system as a graph.}
We model a multi-agent system (MAS) as a directed acyclic graph (DAG) $G=(V,E)$ that captures both agents and their communication patterns. 
Each node $v_i=\langle \texttt{backbone}_i,\ \texttt{role}_i,\ \texttt{state}_i,\ \texttt{plugin}_i \rangle \in V$ corresponds to an agent with four attributes, where \texttt{backbone} denotes the assigned LLM, \texttt{role} denotes the role-specific instruction prompt, \texttt{state} denotes the agent’s accumulated knowledge with interaction history and \texttt{plugin} denotes an optional set of external tools~\cite{toolformer} or APIs~\cite{gptswarm}. 
Each agent $v_i$ receives a prompt $P_i$ and produces a response $\mathcal{R}_i = v_i(P_i)$ according to its attributes.
The prompt $P_i$ is assembled from messages on the incoming edges of the communication topology graph $E \in \{0,1\}^{|V|\times|V|}$, where $E_{ij}=1$ if and only if a message is transmitted from $v_i$ to $v_j$. 

\paragraph{Problem formulation.}
Given a base MAS framework \(M=(G^\ast,\mathcal{B})\), our goal is to maximize task performance while keeping the \emph{expected} token-cost and latency within explicit budgets:
\begin{equation}
\label{eq:goal}
\begin{aligned}
\max_{\theta}\quad 
& \mathbb{E}_{Q\sim\mathcal{D}}\!\left[\mathrm{Perf}\big(F_\theta(M,Q),Q\big)\right] \\
\text{s.t.}\quad
& \mathbb{E}_{Q\sim\mathcal{D}}\!\left[\mathrm{Tok}\big(F_\theta(M,Q),Q\big)\right] \le B_{\mathrm{tok}},\\
& \mathbb{E}_{Q\sim\mathcal{D}}\!\left[\mathrm{Lat}\big(F_\theta(M,Q),Q\big)\right] \le B_{\mathrm{lat}}.
\end{aligned}
\end{equation}
Here \(F_\theta(M,Q)\) denotes the instantiated MAS for query \(Q\); \(F_\theta\) is the configurator parameterized by \(\theta\).
\(\mathrm{Perf}(\cdot)\), \(\mathrm{Tok}(\cdot)\), and \(\mathrm{Lat}(\cdot)\) denote the task performance, token-cost and latency, respectively.

Within the base framework, the template graph \(G^\ast=(T,E^\ast)\) specifies a library of \emph{agent templates} and the admissible edges among them, and \(\mathcal{B}\) is the set of candidate backbones.
An agent template \(\tau=\langle \mathrm{role},\,\mathrm{plugin}\rangle\) serves as an agent prototype that fixes the role and optional tools, while backbone is intentionally left unspecified.
Given a query \(Q\in\mathcal{D}\), the configurator returns a query-specific system
\begin{equation}
\label{eq:definition}
\begin{gathered}
F_\theta(M,Q) \;=\; (V_Q,E_Q),\\
V_Q \;=\; \big\{\langle b_Q(\tau),\,\tau.\mathrm{role},\,\tau.\mathrm{plugin}\rangle:\ \tau\in T_Q\big\},
\end{gathered}
\end{equation}
by selecting a subset \(T_Q\subseteq T\), assigning backbones via the mapping \(b_Q:T_Q\!\to\!\mathcal{B}\), and choosing edges \(E_Q\subseteq E^\ast\). This setup decouples priors and decisions, allowing \textsc{AgentBalance} to optimize performance within the budgets.

\section{Methodology}
\begin{figure*}%[h] % h=here, t=top, b=bottom, p=page, H=exactly here
    \centering
    \includegraphics[width=0.97\textwidth]{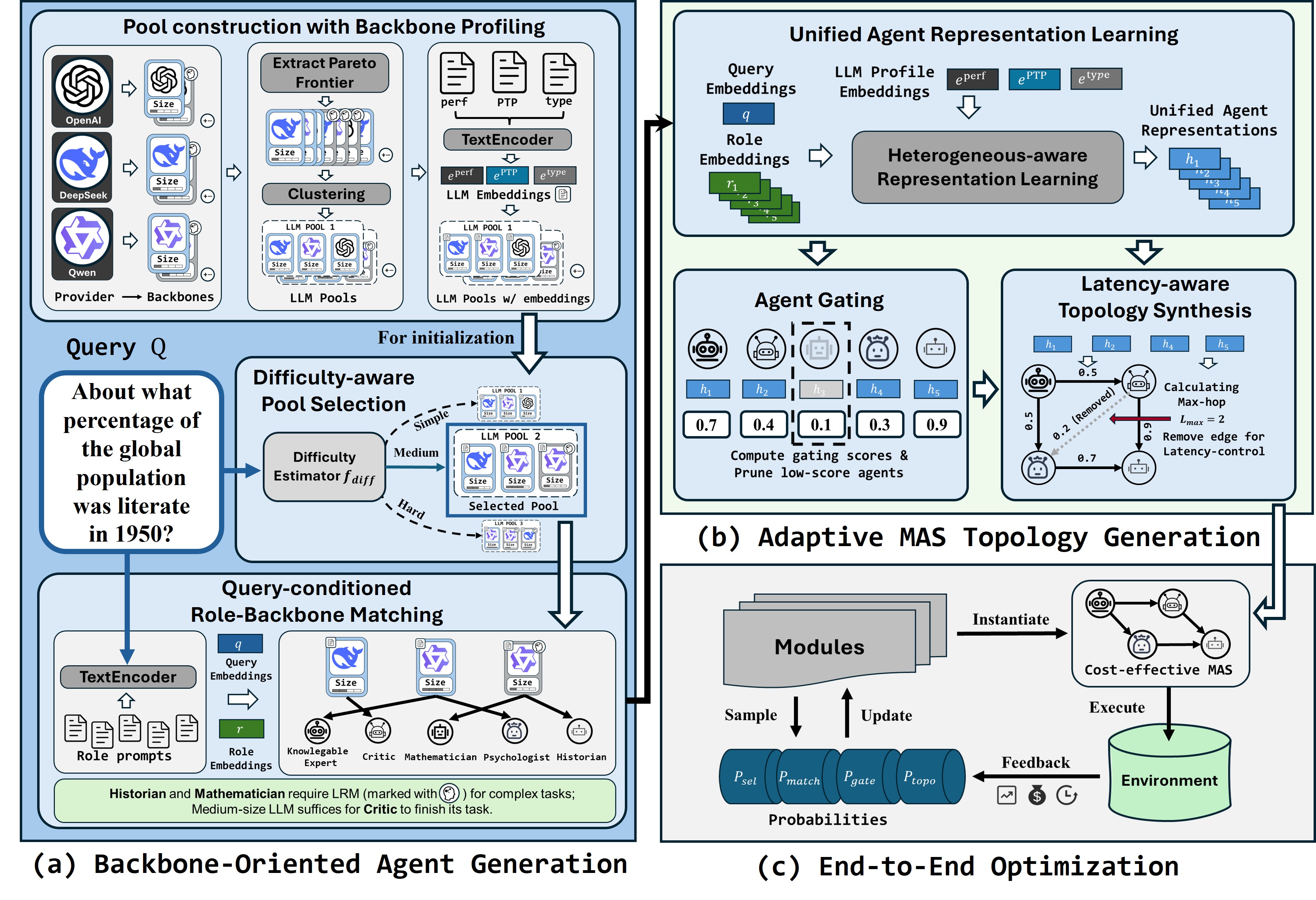}  
    \vspace{-0.4cm}
    \caption{The overall framework of our proposed \textsc{AgentBalance} }
\label{Fig:AB}
    \vspace{-0.4cm}
\end{figure*}

Figure~\ref{Fig:AB} provides an overview of how \textsc{AgentBalance} construct a cost-effective MAS. It first conduct (a) \emph{backbone-oriented agent generation} to produce a set of candidate agents $V_Q$ with heterogeneous backbones for query $Q$. Then (b) \emph{adaptive MAS topology generation} instantiates a latency-aware topology $E_Q$ to form the MAS $G_Q=(V_Q,E_Q)$ based on the candidate agents. (c) \emph{End-to-end optimization} is conducted in the end to optimize the framework under a cost-aware objective to construct cost-effective MAS. We present the high-level algorithm workflow in Algorithm~\ref{alg:agentbalance}.

\subsection{Backbone-Oriented Agent Generation}
\label{sec:CAG}
To construct agents with heterogeneous LLM backbones (varying in size, family, and type), we use a three-stage \emph{backbone-oriented agent generation} pipeline: 
We conduct \emph{backbone-oriented agent generation} in three steps to construct agents with heterogeneous LLM backbones that vary in size, family, and type:
(i) \textit{Pool Construction with Profiling} at initialization; 
(ii) after receiving the query, \textit{Difficulty-Aware Pool Selection} (contributing \(p_{\mathrm{sel}}\)); and 
(iii) \textit{Query-Conditioned Role-Backbone Matching} within the selected pool (contributing \(p_{\mathrm{match}}\)) during execution.
This yields a budget-aligned set of candidate agents \(V_Q\) for the subsequent topology generation.

\subsubsection{Pool Construction with Profiling}
We first filter out a series of non-competitive backbones, then group and profile the remaining LLMs to provide a calibrated basis in the very beginning. This helps with the subsequent pool selection and role-backbone matching.

\textit{Pool Construction with Pareto front.}\quad We construct LLM pools for coherent per-query budget management. For each LLM $m$ from candidate backbones $\mathcal{B}$, we estimate a coarse triple to describe it: $z_m=\langle \mathrm{Perf}_m,\ \mathrm{TokCost}_m,\ \mathrm{Lat}_m\rangle$. $\mathrm{Perf}_m$ is aggregated from benchmarks of $m$; $\mathrm{TokCost}_m$ is the Per-Token Price (PTP) $\times$ token count, with a reasoning-type multiplier applied to LRMs; $\mathrm{Lat}_m$ is a latency proxy derived from the activated parameters of LLM. Since PTP and latency can be distorted by dataset or vendor factors, running a small local calibration can refine these estimates. We retain only LLMs on the 3D Pareto frontier~\citep{pareto} of the triple (\eg \textit{Qwen2.5-72B} is discarded as it is slower and costlier, yet underperforms \textit{Qwen3-32B}). The surviving backbones are then clustered by $k$-medoids over $\phi(m)=\big[\mathrm{Perf}_m,\ \log \mathrm{TokCost}_m,\ \log \mathrm{Lat}_m\big]$ (we use logarithms to group LLMs with similar budget requirement together by ratios). Each cluster defines an LLM pool, and all agents draw from a single selected pool for a single query to enforce a coherent per-query budget during backbone-oriented agent generation. We present more details of this part in Appendix~\ref{appendix:poolconstruction}.

%\paragraph{Backbone Profiling.}
\textit{Backbone Profiling.}\quad For every backbone $m$, we prepare three textual profiles covering (i) performance with benchmark names, (ii) Per-Token Price, and (iii) model type (reasoning or non-reasoning). We use $\mathrm{TextEncoder}$ (a lightweight model that can transform text into representation) encodes them into $e_m^{\mathrm{perf}},\ e_m^{\mathrm{ptp}},\ e_m^{\mathrm{type}}\!\in\!\mathbb{R}^H$. These three LLM embeddings serve as stable descriptors, providing reliable signals of backbone capability and cost.

\subsubsection{Difficulty-Aware Pool Selection}
To improve cost-effectiveness, we allocate the query’s overall budget by estimating difficulty: this module first estimates how hard the query is, then chooses a single resource tier (LLM pool) for agents to match backbones.

\textit{Difficulty Estimation.}\quad
We pretrain a lightweight difficulty estimator $f_{\mathrm{diff}}$, which is a sentence transformer~\citep{mpnet} followed by a small MLP, on the RouterBench dataset~\citep{routerbench}. For supervision, each item is labeled by the fraction of LLMs that solve it. We treat this fraction as a signal of difficulty and train the estimator $f_{\mathrm{diff}}$ (Details of $f_{\mathrm{diff}}$ are in Appendix~\ref{appendix:difficultyestimator}). Only the MLP head of $f_{\mathrm{diff}}$ is updated in the end-to-end optimization. At inference time, the estimator returns a normalized difficulty $d= f_{\mathrm{diff}}(Q) \!\in\![0,1]$. A difficulty offset $\delta$ nudges the operating point toward low-budget or high-budget settings, and a strict upper bound forbids selecting any pool above a user-specified maximum. The final effective difficulty is $d_{\mathrm{eff}}=d+\delta$.

%\paragraph{Pool Assignment.}
\textit{Pool Assignment.}\quad
We map the predicted difficulty to a single resource tier (\ie LLM pool). LLM Pools are ordered from \emph{weak} to \emph{strong} by aggregate performance.
We calculate pool $p$'s representation $e_p$ by averaging the LLM embeddings of its members'. Then we compute cost-aware logits
$\ell_p \;=\; W_P e_p \;-\; \alpha\,\bar{c}_p$, 
where $\alpha$ is a learned cost aversion, and $\bar{c}_p$ is a strictly increasing normalized cost curve over pool indices. 
After masking pools that exceed the strict upper bound, we compute softmax weights over the remaining pools, \(w=\mathrm{softmax}(\mathrm{mask}(\ell))\), and form monotone thresholds by prefix sums, \(\mathrm{thr}_p=\sum_{i\le p} w_i\), which partition \([0,1]\) into contiguous bins. The effective difficulty \(d_{\mathrm{eff}}\) is mapped by a smoothed bucketizer \(G\) (sigmoids centered at \(\{\mathrm{thr}_p\}\)) to a categorical distribution over pools. We then draw a categorical sampler~\citep{categorical} to select the pool $\mathcal{P}_Q$, and the corresponding selection probability is:
\begin{equation}
p_{\mathrm{sel}} \;=\; G_{p^\star}\!\big(d_{\mathrm{eff}};\ \{\mathrm{thr}_p\}_{p=1}^{P-1}).
\label{eq:select}
\end{equation}

\subsubsection{Query-Conditioned Role–Backbone Matching}\quad
Within the selected pool, we assign one backbone to each role by matching to align backbone capability with role needs for cost-effectiveness. 

We first construct LLM representations by aggregating $e_m^{\mathrm{ptp}}$ and $e_m^{\mathrm{type}}$ to capture cost cues, then fuse them with $e_m^{\mathrm{perf}}$ to obtain $u_m=W_u\,[\,e_m^{\mathrm{perf}};\;W_{\mathrm{ct}}[\,e_m^{\mathrm{ptp}};\,e_m^{\mathrm{type}}\,]\,]$.
We encode the query and each agent’s role prompt with the $\mathrm{TextEncoder}$ on the role side to get the query embedding $q$ and the role embedding $r_i$ ($i\in\{1,\dots,|V|\}$). 
% A global pool context $g \;=\; W_g\,\Big(\tfrac{1}{|\mathcal{P}_Q|}\!\sum_{m\in\mathcal{P}_Q}\![\,e_m^{\mathrm{perf}};e_m^{\mathrm{ptp}};e_m^{\mathrm{type}}\,]\Big)$ is formed with the role representation:  $v_i=W_v\,[\,r_i;\,q;\,g\,]$ constructed.
Next, we compute the global pool context
\(g \;=\; W_g\!\left(\tfrac{1}{|\mathcal{P}_Q|}\sum_{m\in\mathcal{P}_Q}[\,e_m^{\mathrm{perf}};e_m^{\mathrm{ptp}};e_m^{\mathrm{type}}\,]\right)\).
Then, we construct the role representation as
\(v_i \;=\; W_v[\,r_i;\,q;\,g\,]\).
Given the role and LLM representations in place, we compute compatibilities use the dot product for per-role distributions, and accumulate the matching probability into the end-to-end objective:
\begin{equation}
p_{\mathrm{match}}=\prod_{i=1}^{|V|}\operatorname{softmax}_m\!\big(\langle v_i,u_m\rangle\big).
\label{eq:match}
\end{equation}
With this, we can assign backbones to agents and instantiate the heterogeneous agents $V_Q$. 
% \vspace{-0.2cm}

\subsection{Adaptive MAS Topology Generation}
\emph{Adaptive MAS topology generation} builds a latency-aware communication topology \(E_Q\) between agents with heterogeneous backbones \(V_Q\), thereby composing the complete MAS: \(G_Q=(V_Q,E_Q)\).
It proceeds in three stages: (i) \textit{Unified Agent Representation Learning} to obtain query- and backbone-aware representations; (ii) \textit{Agent Gating} to remove redundant agents (contributing \(p_{\mathrm{gate}}\)); and (iii) \textit{Latency-Aware Topology Synthesis} to generate a low-latency topology for MAS (contributing \(p_{\mathrm{topo}}\)).
\label{sec:HMASC}

\subsubsection{Unified Agent Representation Learning}\quad
Agents obtained from \emph{backbone-oriented agent generation} are equipped with heterogeneous backbones, whose size, family and type may differ across roles, and these backbones also switch across queries. To handle this strong heterogeneity in determining agent communication patterns, we conduct a representation learning and project such heterogeneity into a \emph{unified, query- and backbone-conditioned} representation for agents to capture role semantics together with backbone capability, query information and cost cues. This unified agent representation serves as the input to agent gating and latency-aware topology synthesis in the following parts.

Specifically, for role $i$ with selected backbone $m_i$, we first form prototypes with query and role embeddings by $q^{h}=W_q q$ and $r_i^{h}=W_r r_i$. 
We then map the three LLM embeddings through a shared head $W_{\ell}$: 
$d_i=W_{\ell} e^{\mathrm{perf}}_{m_i}$, $c_i=W_{\ell} e^{\mathrm{ptp}}_{m_i}$, and $t_i=W_{\ell} e^{\mathrm{type}}_{m_i}$. 
Next, we compute a query-conditioned context with 
$\mathbf{q}_i = W_{\mathbf{q}}\,[\,r_i^{h};\,q^{h}\,]$, 
$\mathbf{k}_i = W_{\mathbf{k}}\,[\,d_i;\,c_i;\,t_i\,]$, 
$\mathbf{v}_i = W_{\mathbf{v}}\,[\,d_i;\,c_i;\,t_i\,]$, 
and obtain the unified agent representation by
\begin{equation}
h_i \;=\; r_i^{h} \;+\; \gamma\,W_{\mathrm{ctx}}\,\mathrm{attn}(\mathbf{q}_i,\mathbf{k}_i,\mathbf{v}_i),
\end{equation}
where $\mathrm{attn}(\mathbf{q},\mathbf{k},\mathbf{v})$ denotes scaled dot-product attention~\citep{attn} and $\gamma$ is a trainable scale. 
This design injects a lightweight \emph{query- and backbone-aware} context into the agent representation $\{h_i\}$, which facilitates  \emph{Agent Gating} and \emph{Latency-Aware Topology Synthesis}.

\subsubsection{Agent Gating}\quad
Not all agents are necessary for a given query (\eg a \emph{psychologist} agent for a math problem). To improve cost-effectiveness for MAS construction, this module prunes redundant agents while preserving a compact, competent set for answering a question. Given the unified agent representations $\{h_i\}_{i=1}^{|V_Q|}$, we at first form a global context $\bar h=\tfrac{1}{|V_Q|}\sum_i h_i$ and then compute agent keep probabilities with a single scoring step $p_i \;=\; \sigma\!\big(W_g[\,h_i;\bar h\,]\big)$.
We adopt Bernoulli sampling for the keep decision and use a Gumbel-Sigmoid (Concrete) relaxation~\citep{gumbel} during training for differentiability. Let $g_i\in\{0,1\}$ denote the sampled decision, then the gating probability for agent gating is
\begin{equation}
p_{\mathrm{gate}}=\prod_{i=1}^{|V|}p_i^{\,g_i}\,(1-p_i)^{\,1-g_i}.
\end{equation}
We ensure at least two active agents by activating the highest-$p_i$ items if needed; the retained set $V_{Q}^{\mathrm{gated}}=\{v_i\in V_{Q}: i\in (g_i=1)\} $ is then passed to the topology generator.

\subsubsection{Latency-Aware Topology Synthesis}\quad
Excessive inter-agent communication may degrade accuracy and inflate latency. We therefore synthesize a latency-aware communication topology over the retained agents while controlling end-to-end delay. Using the unified agent representations $\{h_i\}_{i=1}^{N}$ ($N\, = \,|V_{Q}^{\mathrm{gated}}|$), we compute edge probabilities: $p_{ij}=\sigma(\langle W_a h_i,\; W_a h_j\rangle)$. 
Then, edges are sampled with a Gumbel-Sigmoid (Concrete) relaxation to obtain the stochastic adjacency $E_Q=E_{ij}$  ($i,j\in (g_i=1)$). The probability of topology synthesis is
\begin{equation}
p_{\mathrm{topo}}=\!\!\!\!\prod_{1\le i<j\le N}\!\! p_{ij}^{\,E_{ij}}\,(1-p_{ij})^{\,1-E_{ij}}.
\end{equation}

Because long reasoning chains increase inference latency, we infer a \emph{hop limit} to constrain path depth.
We first predict a distribution over hop increments \(k\in\{1,\ldots,N-1\}\) from the pooled context \(\bar h=\tfrac{1}{N}\sum_i h_i\): let \(\pi_{\mathrm{hop}}=\mathrm{softmax}(W_L\bar h)\in\Delta^{N-2}\) denote the Gumbel-Softmax simplex.
We then set the \emph{hop limit} to a one-hop baseline plus the expected increment,
\begin{equation}
L_{\max}=1+\sum_{k=1}^{N-1} k\,\pi_{\mathrm{hop}}(k) .
\end{equation}
Let \(\ell(E_Q)\) be the longest-path length on the sampled edges \(E_Q\).
If \(\ell(E_Q)>L_{\max}\), we iteratively remove the lowest-probability edge on the current critical path until the constraint is met.
During training, we use a hop-length penalty \(\mathrm{Pen}_{\mathrm{len}}=\operatorname{ReLU}\!\big(\ell(E_Q)-L_{\max}\big)\) to learn  \(L_{\max}\) and encourage feasible topologies.

\subsection{End-to-end Optimization}
\label{sec:Optimization}
We optimize the pipeline with a {Lagrangian surrogate} that balances task performance with token-cost and latency. 
For a query $Q$, the instantiated MAS is $G_{Q}=F_\theta(M,Q)$ and the pre-query reward is
\begin{equation}
\begin{gathered}
R(G_{Q},Q)\;=\;\mathrm{Perf}\!\big(F_\theta(M,Q),Q\big)\\-\;\lambda_{\mathrm{tok}}\>\mathrm{Tok}\!\big(F_\theta(M,Q),Q\big)\;-\;\lambda_{\mathrm{lat}}\>\mathrm{Lat}\!\big(F_\theta(M,Q),Q\big).   
\end{gathered}
\label{eq:reward_e2e}
\end{equation}
$\lambda_{\mathrm{tok}},\lambda_{\mathrm{lat}}\!\ge\!0$ are two hyperparameters that index operating points on the trade-off among performance, token-cost and latency.

Following our backbone-then-topology design, we first perform backbone-oriented agent generation to narrow the search space and then, conditioned on the selected backbones, conduct adaptive MAS topology generation to refine the configuration. For each query, the configurator samples a discrete decision tuple \(\mathcal{D}=\{p,\{m_i\},\{g_i\},E_Q\}\). After executing $G_{Q}$ to obtain $R$, we can aggregate the policy signal through the following factorization:
\begin{equation}
p_\theta(\mathcal{D}\mid Q)\;=\; \underbrace{\textstyle p_{\mathrm{sel}}}_{\text{pool selection}}\cdot \underbrace{\textstyle p_{\mathrm{match}}}_{\text{backbone}\to\text{role}}\cdot \underbrace{\textstyle p_{\mathrm{gate}}}_{\text{gating}}\cdot \underbrace{\textstyle p_{\mathrm{topo}}}_{\text{topology}}.
\label{eq:prob_factor_e2e}
\end{equation}
With this, we can optimize a single policy-gradient objective with a hop-length regularizer:
\begin{equation}
\mathcal{L}\;=\;-\;R(G_Q,Q)\,\log p_\theta(\mathcal{D}\mid Q)\;+\;\lambda_{\mathrm{len}}\>\mathrm{Pen}_{\mathrm{len}}\!\big(E_Q,L_{\max}\big),
\label{eq:pg_e2e}
\end{equation}
where $\lambda_{\mathrm{len}}\geq0$ balances hop-length against reward. 
This objective improves cost-effectiveness by increasing the likelihood of correct solutions while regularizing hop-length, thereby curbing token usage and latency.
Then, we follow standard practice in MAS design~\citep{gptswarm} and apply policy gradient~\citep{policy_gradient} to minimize $\mathcal{L}$;

\begin{table*}[t]
\centering
\setlength{\tabcolsep}{3.5pt}
\renewcommand{\arraystretch}{1.03}
\caption{Performance comparison with baselines under token-cost and latency budgets (\(P@T_{1..4}\), \(P@L_{1..4}\); details in Appendix~\ref{appendix:budgets}). AUCs of the performance–budget curves are also reported. All metrics are higher-is-better (↑); the best per dataset in \textbf{bold}.}
\vspace{-0.4cm}
\begin{tabular}{l l *{5}{c} *{5}{c}}
\toprule
& & \multicolumn{5}{c}{\textbf{Under Token-Cost Budgets}} & \multicolumn{5}{c}{\textbf{Under Latency Budgets}} \\
\cmidrule(lr){3-7}\cmidrule(lr){8-12}
\textbf{Dataset} & \textbf{Method} & $P@T_1$ & $P@T_2$ & $P@T_3$ & $P@T_4$ & {AUC$_{tok}$} & $P@L_1$ & $P@L_2$ & $P@L_3$ & $P@L_4$ & {AUC$_{lat}$} \\
\midrule
% ===== MMLU =====
\multirow{5}{*}{MMLU}
& AgentPrune   & 59.68 & 79.21 & 83.48 & 86.77 & 1.269 & 32.58 & 71.25 & 82.21 & 87.52 & 237.7 \\
& AgentDropout & 61.45 & 80.22 & 83.20 & 85.98 & 1.262 & 39.50 & 81.92 & 82.79 & 84.93 & 238.2 \\
& G-Designer   & 61.80 & 74.03 & 83.59 & 86.40 & 1.261 & 59.69 & 74.34 & 82.11 & 86.61 & 243.2 \\
& MasRouter    & 15.18 & 28.19 & 69.36 & 82.39 & 1.117 & 51.59 & 81.13 & 83.15 & 84.97 & 239.2 \\
& \textsc{AgentBalance}
               & \textbf{71.90} & \textbf{83.66} & \textbf{85.62} & \textbf{88.02} & \textbf{1.297}
               & \textbf{71.90} & \textbf{83.66} & \textbf{85.62} & \textbf{88.02} & \textbf{250.0} \\
\midrule
% ===== HumanEval =====
\multirow{5}{*}{HumanEval}
& AgentPrune   & 86.56 & 87.21 & 89.26 & 94.64 & 1.868 & 68.49 & 86.42 & 90.95 & 92.76 & 470.6 \\
& AgentDropout & 87.23 & 87.95 & 91.06 & 95.01 & 1.876 & 70.42 & 88.95 & 91.07 & 93.86 & 471.0 \\
& G-Designer   & 86.90 & 87.64 & 90.73 & \textbf{95.52} & \textbf{1.885} & 83.81 & 88.35 & 91.42 & 93.51 & 474.9 \\
& MasRouter    & 12.68 & 15.90 & 51.16 & 90.86 & 1.724 & 16.74 & 21.47 & 77.89 & 88.83 & 440.9 \\
& \textsc{AgentBalance}
               & \textbf{87.94} & \textbf{89.13} & \textbf{92.20} & 95.46 & 1.880
               & \textbf{87.94} & \textbf{89.13} & \textbf{92.20} & \textbf{95.46} & \textbf{476.5} \\
\midrule
% ===== MATH =====
\multirow{5}{*}{MATH}
& AgentPrune   & 63.96 & 72.09 & 75.74 & 78.98 & 3.515 & 60.28 & 66.35 & 72.12 & 75.49 & 592.9 \\
& AgentDropout & 62.75 & 72.17 & 74.91 & 78.05 & 3.471 & 60.19 & 68.21 & 72.27 & 75.52 & 588.5 \\
& G-Designer   & 63.05 & 71.26 & 74.80 & 77.45 & 3.450 & 59.43 & 64.73 & 71.24 & 74.26 & 581.9 \\
& MasRouter    & 19.26 & 39.46 & 67.81 & 74.29 & 3.259 & 36.88 & 65.38 & 69.92 & 75.63 & 565.4 \\
& \textsc{AgentBalance}
               & \textbf{66.46} & \textbf{73.33} & \textbf{75.83} & \textbf{79.38} & \textbf{3.523}
               & \textbf{66.46} & \textbf{73.33} & \textbf{75.83} & \textbf{79.38} & \textbf{602.4} \\
\bottomrule
\end{tabular}
\vspace{-0.4cm}
\label{tab:ab_one_table}
\end{table*}

\section{Experiments}
\subsection{Experimental Setup}

\paragraph{Benchmarks}
%\textit{Benchmarks}\quad
We evaluate \textsc{AgentBalance} on benchmarks from three domains: \textbf{(1) MMLU} for general knowledge; \textbf{(2) MATH} for formal mathematical problem solving and reasoning; and \textbf{(3) HumanEval} for code generation. 
These benchmarks span multiple domains and reasoning skills, providing a comprehensive assessment of cross-domain effectiveness.

\paragraph{Baselines and LLM Backbones}
We compare against cost-effective MAS baselines, including the \emph{single-LLM MAS} methods AgentPrune~\citep{AgentPrune}, AgentDropout~\citep{agentdropout}, and G-Designer~\citep{G-Designer}, as well as the \emph{multi-LLM MAS} baseline MasRouter~\citep{masrouter}.
All baselines are constructed based on Complete Graph~\citep{macnet}.
Candidate LLMs consist of 14 backbone models from Qwen and DeepSeek, selected to evaluate the cost-effectiveness of \textsc{AgentBalance}. Detailed configuration of LLMs is in Appendix~\ref{appendix:candidate_LLMs}.

\paragraph{Implementation}
We set the learning rate to $0.1$ and fix the LLM temperature at $0$. We instantiate four LLM pools and employs an MPNet $\mathrm{TextEncoder}$~\citep{mpnet} in \textsc{AgentBalance}. 
For data efficiency and fair comparison, each method is trained with $20$ samples on HumanEval and $40$ on the others datasets, and we report averages over multiple independent runs.
To characterize attainable trade-offs, single-LLM baselines are evaluated using all these candidate backbones and their \emph{envelope} defines the cost-effectiveness frontier; for MasRouter and \textsc{AgentBalance}, we sweep hyperparameters to cover a range of token-cost and latency budgets.More implementation details are presented in Appendix~\ref{appendix:implementation}. 

\paragraph{Metrics}
We adopt \emph{Performance-at-Budget}~\citep{pk} for both token-cost and latency. 
\(P@T\) denotes the task performance measured {subject to a token-cost budget}, and \(P@L\) analogously under a latency budget. The task performance is \emph{accuracy} on \textsc{MMLU} and \textsc{MATH}, and \emph{Pass@1} on \textsc{HumanEval}. 
To capture overall behavior across budgets, we also report the AUC of the performance–token-cost and performance–latency curves. 
Details of the metrics are in Appendix~\ref{appendix:metrics}.

\subsection{Main results}
Table~\ref{tab:ab_one_table} presents results of \textsc{AgentBalance} and baselines under token-cost and latency budgets.
Specifically, \(P@T_{1..4}\) and \(P@L_{1..4}\) evaluate performance at four monotonically increasing budget points for token-cost and latency, respectively; these points are chosen to cover a broad operating range from low to high budgets (exact values are given in Appendix~\ref{appendix:budgets}). In addition to these point-wise measurements, we report the area under the performance–token-cost and performance–latency curves (AUC), which summarizes overall behavior across the entire budget spectrum.

We verify that \textsc{AgentBalance} is:
\textbf{(1) Token-cost efficiency}: it achieves higher performance at matched token budgets (\(P@T_{1..4}\)); \eg at the tightest budget \(P@T_1\) on \textsc{MMLU}, it reaches \(\mathbf{71.90}\) vs.\ 61.80 (G-Designer) and 61.45 (AgentDropout). It also remains high at \(P@T_4\): \(\mathbf{88.02}\) vs.\ 86.77 (AgentPrune), confirming strong performance across token-cost budgets.
\textbf{(2) Latency efficiency}: it attains strong performance under end-to-end latency constraints across budgets (\(P@L_{1..4}\)); \eg on \textsc{MATH} \(P@L_1=\mathbf{66.46}\) vs.\ 60.28 (G-Designer), and on \textsc{HumanEval} \(P@L_4=\mathbf{95.46}\) vs.\ 92.76 (AgentPrune). These results highlight the effectiveness of the latency-aware components in \textsc{AgentBalance}.
\textbf{(3) Global cost-effectiveness}: across datasets, \textsc{AgentBalance} yields consistent gains and higher AUCs, indicating balanced improvements beyond a single operating point. While it is slightly below G-Designer on \textsc{HumanEval} at the highest token-cost budget, its strong performance at low and mid budgets keeps it competitive overall.
In contrast, single-LLM baselines often underperform as they overlook backbone choices in MAS, and \textsc{MasRouter}'s topology-first design cannot adequately accommodate heterogeneous LLMs with widely disparate capabilities or their induced topology effects, leading to inconsistent performance across budgets and weaker results in low-budget regimes.

\begin{table*}[t]
\centering
% \small
\setlength{\tabcolsep}{6pt}
\renewcommand{\arraystretch}{1.05}
\caption{Plug-in effectiveness on the MMLU dataset under token-cost and latency budgets (\(P@T_{1..4}\), \(P@L_{1..4}\); details in Appendix~\ref{appendix:budgets}). AUCs of the performance–budget curves are also reported. All metrics are higher-is-better (↑); the best per dataset in \textbf{bold}.}
% \caption{Plug-in effectiveness on MMLU dataset at token-cost (\(P@T_{1..4}\)) and latency (\(P@L_{1..4}\)) budgets (details in Appendix~\ref{appendix:budgets}). We also report AUC of the performance--budget curves for both token-cost and latency. Best method per dataset is \textbf{bold}.}
\vspace{-0.4cm}
\begin{tabular}{l *{5}{c} *{5}{c}}
\toprule
& \multicolumn{5}{c}{\textbf{Under Token-Cost Budgets}} & \multicolumn{5}{c}{\textbf{Under Latency Budgets}} \\
\cmidrule(lr){2-6}\cmidrule(lr){7-11}
Method & $P@T_1$ & $P@T_2$ & $P@T_3$ & $P@T_4$ & AUC$_{tok}$ & $P@L_1$ & $P@L_2$ & $P@L_3$ & $P@L_4$ & AUC$_{lat}$ \\
\midrule
Layered Graph~\citep{macnet} & 69.26 & 79.01 & 85.69 & 86.49 & 1.414 & 45.38 & 59.79 & 86.41 & 87.35 & 165.93 \\
\textbf{+ \textsc{AgentBalance}} & \textbf{73.20} & \textbf{80.83} & \textbf{86.71} & \textbf{88.67} & \textbf{1.430} & \textbf{73.20} & \textbf{80.83} & \textbf{86.71} & \textbf{88.67} & \textbf{173.61} \\
\midrule
AutoGen~\citep{autogen} & 69.20 & 81.48 & 85.56 & 87.13 & 1.449 & 58.97 & 76.58 & 86.04 & 87.34 & 155.13 \\
\textbf{+ \textsc{AgentBalance}} & \textbf{72.55} & \textbf{82.35} & \textbf{86.40} & \textbf{89.76} & \textbf{1.473} & \textbf{72.55} & \textbf{82.35} & \textbf{86.40} & \textbf{89.76} & \textbf{160.55} \\
\bottomrule
\end{tabular}
\label{tab:plug_in_effect}
\vspace{-0.4cm}
\end{table*}

\subsection{Plug-in to Existing MAS}
We evaluate \textsc{AgentBalance} as a framework-agnostic plug-and-play module by inserting it into two MAS frameworks, AutoGen~\citep{autogen} and Layered Graph~\citep{macnet}, replacing their backbone-topology configuration with \textsc{AgentBalance}. As shown in Table~\ref{tab:plug_in_effect}, \textsc{AgentBalance} consistently performs better given token-cost and latency budgets. At tight budgets, $P@T1$ improves from 69.26$\rightarrow$73.20 on Layered Graph and 69.20$\rightarrow$72.55 on AutoGen, and the corresponding latency points $P@L1$ rise from 45.38$\rightarrow$73.20 and 58.97$\rightarrow$72.55. The performance gains persist at higher budgets, accompanied by AUC increases on both axes. These results indicate that \textsc{AgentBalance} can serve as a cost-effective plug-in for existing MAS, fostering a practical, budget-aware deployment.

\begin{figure*}%[h] % h=here, t=top, b=bottom, p=page, H=exactly here
    \centering
    \includegraphics[width=0.9\textwidth]{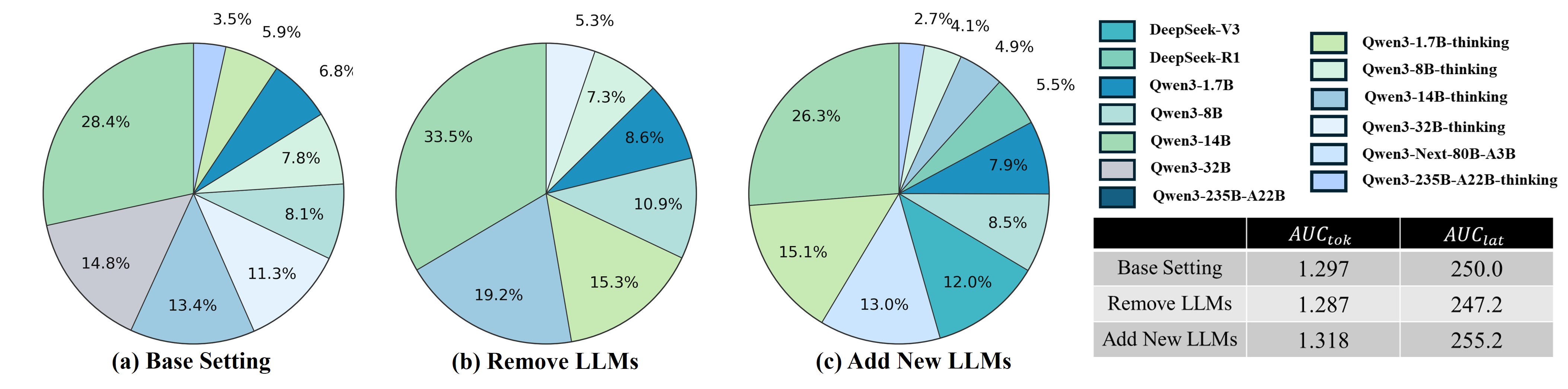}  %AB_Intro.png
    \vspace{-0.3cm}
    \caption{Inductive ability analysis of \textsc{AgentBalance} on MMLU. (a)-(c) show the proportions of LLM invocations under different LLM configurations. AUC\textsubscript{tok} and AUC\textsubscript{lat} are listed on the right.}
    \vspace{-0.4cm}
    \label{fig:inductive}
\end{figure*}

\subsection{Inductive Ability Analysis}
We evaluate whether \textsc{AgentBalance} generalizes to new LLM configurations without retraining. Two interventions are considered: (i) removing two backbones (thinking and non-thinking versions of Qwen3-235B-A22B) from the candidate backbones; and (ii) adding three more cost-effective backbones (Qwen3-Next-80B-A3B, DeepSeek-V3, DeepSeek-R1). As shown in Figure~\ref{fig:inductive}, \textsc{AgentBalance} reconfigures the MAS using the updated LLMs and performs inference with models not seen during training. When backbones are removed, the performance--budget frontier shifts downward and AUC decreases, yet the method remains superior to baselines. When new backbones are added, the frontier improves and AUC increases markedly, demonstrating strong inductive ability and rapid adaptation to new models without additional training resources.
\label{sec:inductive}

\subsection{Framework Analysis}

\begin{table}[t]%[11]{r}{0.48\columnwidth}
% \vspace{-12pt}
\captionsetup{font=small,justification=raggedright,singlelinecheck=false}
\caption{Ablation on \textsc{AgentBalance}. We conduct evaluations with three metrics: Perf (\%) (↑), Token-cost (USD) (↓) Latency (s) (↓).}
\vspace{-0.3cm}
\setlength{\tabcolsep}{4pt}\renewcommand{\arraystretch}{1.05}
\begin{tabular}{l r r r r r r}
\toprule
& \multicolumn{3}{c}{\textbf{MATH}} & \multicolumn{3}{c}{\textbf{MMLU}} \\
\cmidrule(lr){2-4}\cmidrule(lr){5-7}
\textbf{Method} & Perf↑ & Tok↓ & Lat↓ & Perf↑ & Tok↓ & Lat↓ \\
\midrule
\textsc{AgentBalance} & 0.758 & 0.496 & 99.3 & 0.856 & 0.315 & 63.5 \\
Variant (1) & 0.733 & 0.838 & 169.4 & 0.824 & 0.394 & 69.7 \\
Variant (2) & 0.713 & 0.295 & 53.5 & 0.800 & 0.202 & 50.1 \\
Variant (3) & 0.746 & 0.871 & 141.1 & 0.856 & 0.351 & 74.6 \\
Variant (4) & 0.754 & 0.548 & 114.3 & 0.850 & 0.418 & 77.6 \\
Variant (5) & 0.758 & 0.815 & 141.1 & 0.861 & 0.564 & 99.3 \\
Variant (6) & 0.744 & 0.626 & 171.3 & 0.850 & 0.319 & 88.6 \\
\bottomrule
\end{tabular}
\label{tab:ablation}
    \vspace{-0.4cm}
\end{table}

\paragraph{Ablation Study}
We consider six variants for the ablation study: (1) constructing random LLM pools, (2) random pool selection, (3) random role-backbone matching, (4) using role embeddings only in adaptive MAS topology generation, (5) removing agent gating and (6) apply dense communication topology. 
The results are present in Table~\ref{tab:ablation}:
(1) Constructing random pools or (2) a random pool selection causes a substantial drop in performance. (3) A random role–backbone matching and (4) ignoring backbone and query information in topology generation markedly reduce cost-effectiveness. Furthermore, (5) eliminating agent gating or (6) applying a dense topology leads to significant latency increases. These results verify that every module in our design contributes to the overall performance--token-cost--latency trade-off. 

\begin{figure}%[h] % h=here, t=top, b=bottom, p=page, H=exactly 
    \centering
    \includegraphics[width=0.48\textwidth]{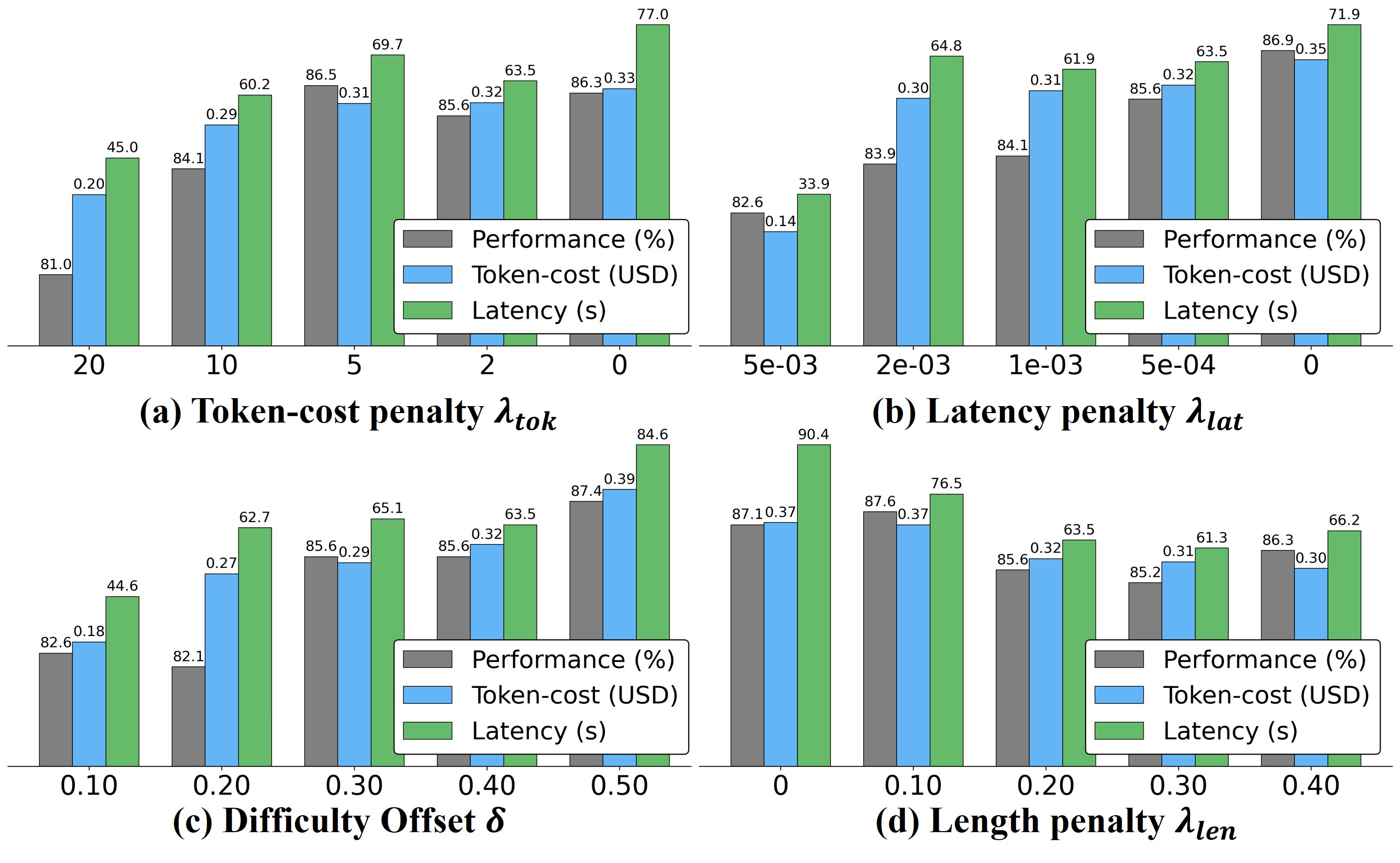} 
    \vspace{-0.4cm}
    \caption{Hyperparameter analysis of \textsc{AgentBalance}.}
    \label{fig:hparam_analysis}
    %\vspace{-0.4cm}
\end{figure}

\paragraph{Sensitivity Analysis}
We present the effect four hyperparameters of \textsc{AgentBalance} in Figure~\ref{fig:hparam_analysis}(a)--(d), and we highlight the key findings below.
\textbf{(a) Token-cost weight $\lambda_{\text{tok}}$:}
As $\lambda_{\text{tok}}$ increases, decisions shift toward cheaper configurations; correspondingly, token-cost decreases, with latency and accuracy declining slightly (as small backbones are encouraged in cost-effective agent generation). This hyperparameter offers a balanced trade-off to achieve a specific budget.
\textbf{(b) Latency weight $\lambda_{\text{lat}}$:}
With higher $\lambda_{\text{lat}}$, slow configurations are discouraged, which typically lowers latency. In return, accuracy may drop, whereas relaxing this weight yields the opposite behavior.
\textbf{(c) Difficulty offset $\delta$:}
As $\delta$ increases, the selector is more likely to choose heavier pools; consequently, accuracy tends to improve mildly, whereas token-cost and latency rise.
\textbf{(d) Length penalty $\lambda_{\text{len}}$:}
By penalizing long communication chains, larger $\lambda_{\text{len}}$ steadily reduces latency.

\begin{figure}%[h] % h=here, t=top, b=bottom, p=page, H=exactly here
    \centering
    \includegraphics[width=0.45\textwidth]{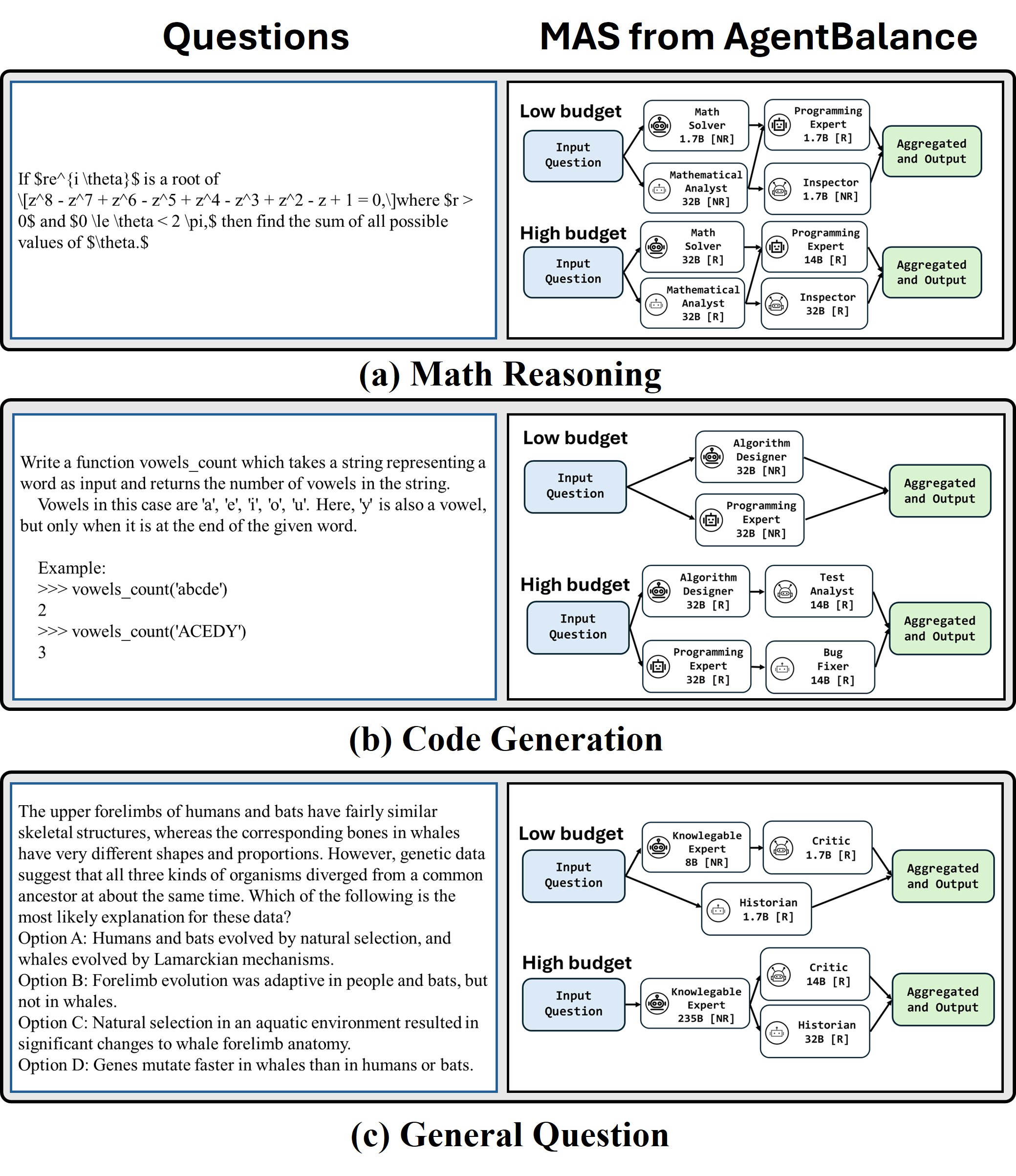}
    %\vspace{-0.4cm}
    %AB_Intro.png
    \caption{Case study of \textsc{AgentBalance}. Backbones are from the Qwen3 family, with [R] for reasoning models and [NR] for non-reasoning models.}
    \label{fig:case}
    \vspace{-0.4cm}
\end{figure}

\subsection{Case Study}
We conduct a case study of \textsc{AgentBalance} on three tasks with different resource budgets. We observe that \textsc{AgentBalance} (i) \emph{performs diverse backbone assignment and topology generation}: as shown in the high budget setting of (a), it places 32B backbones on the other agents while attaching a lighter \emph{Inspector}. It also generates a layered topology in (a) and a parallel topology in (b), respectively, demonstrating its capability to adapt to diverse environments; (ii) \emph{instantiates the MAS according to budget}: as shown in (b), the low-budget configuration keeps a lean two-agent pipeline without an LRM, whereas the high-budget configuration adds verification agents and enables LRMs as backbones; (iii) \emph{exhibits structural stability across budgets}: the core task structures remain stable while capacity increases. \eg in general QA the three selected agents are retained across budgets, indicating that \textsc{AgentBalance} learns task-level structural information to construct the MAS.

\section{Related Work}
\textbf{LLM-based Multi-Agent Systems}\quad
Theoretically grounded in the society-of-mind perspective~\citep{SocietyOfMind}, LLM-based multi-agent systems (MAS) coordinate specialized role prompts, tool interactions, and inter-agent collaboration to deliver performance that often surpasses single-LLM systems~\citep{react,reflexion,agentverse,multiagentdebate,toolformer}. Moving beyond pre-defined architectures, recent studies begin to define MAS as graphs and {learn} the communication topology between agents to strengthen coordination, revealing that {who talks to whom and when} substantially impacts system performance~\citep{gptswarm,AgentPrune,G-Designer,MaAS}. Another line of work defines agent interactions via {workflows}~\citep{aflow,flow,agentsquare}, where stage-wise task graphs and tool pipelines are specified up front, offering modularity but typically less adaptivity than topology-learning approaches.  While the community emphasizes that communication patterns of agents should be {optimized as a first-class design variable} rather than treated as fixed~\citep{MAS_topo, InfoPropagation}, most MAS implicitly assume the use of strong homogeneous LLMs and thus overlook the substantial influence of {backbone choice} on system behavior. Even MAS that employ multiple LLMs~\citep{X-MAS,Heterogeneous_swarms} remain primarily accuracy-oriented: they select backbones with similar capabilities per task and do not focus on collaboration between LLMs with disparate capabilities at different budgets.\\
\textbf{Cost-effectiveness in MAS}\quad
The high performance of MAS is tightly coupled with high cost due to frequent LLM invocation, long inference pipeline, intensive inter-agent communication, and a reliance on strong LLM backbones (including LRMs). Against this backdrop, the community starts focusing on the cost-effectiveness issue and explores token-cost reduction methods. AgentPrune~\cite{AgentPrune} classifies inter-agent communication into spatial/temporal channels and prunes redundant messages, while G-Designer~\cite{G-Designer} uses a GNN to synthesize sparse topologies, but both of them chiefly cut \emph{prompt} tokens. Because decoding of \emph{completion} tokens dominates compute and billing in modern LLM inference~\citep{tngtech_prefill_decode_blog} and prefill/input tokens are typically priced lower~\citep{openai_pricing,anthropic_pricing}, pruning prompts alone misses the main cost driver, thereby limiting the effectiveness of these methods.
Building upon AgentPrune, AgentDropout~\cite{agentdropout} additionally removes agents that contribute less, yet its pruning is not query-adaptive and thus can degrade system performance. MasRouter~\citep{masrouter} introduces multi-LLM routing during MAS construction and considers backbone choice with cost awareness. However, its design remains topology-first and does not model how backbone choices reshape the communication topology. Moreover, it neither addresses large performance disparities across candidate backbones nor provides role-conditioned backbone assignment, which limits cost-effectiveness especially at low-budget regimes.

\section{Conclusion}
This paper presents \textsc{AgentBalance}, a framework for constructing cost-effective MAS with backbone-then-topology design under token-cost and latency budgets. \textsc{AgentBalance} first conducts backbone-oriented agent generation to construct cost-effective candidate agents by pooling heterogeneous LLMs, selecting LLM pools, and performing role-backbone matching. It then conducts Adaptive MAS Topology Generation via agent representation learning, gating redundant agents, and synthesizing latency-aware communication topology. Experiments demonstrate that \textsc{AgentBalance} achieves consistent performance gains across different token-cost and latency budgets in different regimes. In addition, it can serve as a plug-in to enhance existing MAS and generalize to unseen LLMs with high performance.
These results provide actionable guidance for deploying MAS in large-scale AI-powered web applications under explicit token-cost and latency budgets, enabling predictable latency, scalable operation, and cost control at web scale.

\clearpage
\bibliographystyle{ACM-Reference-Format}
\bibliography{bibfile}

\appendix

\section{Candidate LLMs}
\label{appendix:candidate_LLMs}
We summarize the candidate backbones and their per-token prices (PTP) in Table~\ref{tab:LLM-price}. Most models are accessed via the Siliconflow online API~\cite{Siliconflow} where two smaller LLMs (Qwen3-8B and Qwen3-1.7B) are served locally using \texttt{vLLM}~\citep{vLLM}. To ensure consistent budgeting, we price them according to the PTP on Siliconflow. 

Our selected candidate LLMs span parameter scales from \(\,1.7\mathrm{B}\) to \(235\mathrm{B}\) and include both reasoning and non-reasoning models. We also include a Qwen2.5 variant (Qwen2.5-72B-Instruct) for completeness, though it is dominated under our Pareto-based pool construction and is therefore never selected at runtime. 

For transferability analysis, we include \emph{Qwen3-Next-80B-A3B}, \emph{DeepSeek-V3}, and \emph{DeepSeek-R1}. Qwen3-Next-80B-A3B ~\citep{qwen3_next} is the new generation of the Qwen3 series and it exhibits better cost-effectiveness. The Deepseek model are selected as a new family of LLMs to demonstrate the effectiveness of LLM profiling, and we price them at \textbf{half} of the PTP on Siliconflow on DeepSeek models to simulate a cost-effective provider. Overall, this backbone set allows us to evaluate \textsc{AgentBalance} across diverse model sizes, types, and vendors.

\begin{table}[t]
\centering
\caption{LLM backbones and unit prices used in \textsc{AgentBalance}. PTPs are in CNY per million tokens. We employ  \emph{both} non-reasoning and reasoning types for most Qwen3 models, where their PTP is identical across types but LRMs typically yield longer completions.}
\vspace{-0.3cm}
\label{tab:LLM-price}
\small
\setlength{\tabcolsep}{6pt}
\renewcommand{\arraystretch}{1.08}
\begin{tabular}{l l r r}
\toprule
\textbf{Model} & \textbf{Type} & \textbf{Input PTP} & \textbf{Output PTP} \\
\midrule
\multicolumn{4}{l}{\textbf{Primary LLM Backbones}} \\
\addlinespace[2pt]
Qwen3-235B-A22B & Both & 2.5  & 10.0 \\
Qwen3-32B       & Both & 1.0  & 4.0  \\
Qwen3-14B       & Both & 0.5  & 2.0  \\
Qwen3-8B        & Both & 0.25 & 1.0  \\
Qwen3-1.7B      & Both & 0.1  & 0.4 \\
Qwen2.5-72B-Instruct & non-reasoning & 4.13 & 4.13 \\
\midrule
\multicolumn{4}{l}{\textbf{Transferability Experiment Only}} \\
\addlinespace[2pt]
Qwen3-Next-80B-A3B       & non-reasoning & 1.0  & 4.0  \\
DeepSeek-V3     & non-reasoning & 1.0  & 4.0 \\
DeepSeek-R1     & reasoning    & 2.0  & 8.0 \\
\bottomrule
\end{tabular}
\vspace{-0.6cm}
\end{table}

\section{Details of Pool Construction and Profiling}
\label{appendix:poolconstruction}
We provide more details for \emph{Pool Construction and Profiling} here.

\textbf{Pool construction.}\quad
We aggregate LLMs with similar \emph{performance}, \emph{token-cost}, and \emph{latency} ($\mathbf{z}_m=\langle \mathrm{Perf}_m,\mathrm{TokCost}_m,\mathrm{Lat}_m\rangle$) into LLM pools for better resource control. We show how we construct the triple, then provide more details in pool construction.

We initially incorporate the benchmark performance of corresponding tasks into the profile. Although per-token-prices (PTP) are available, the token-cost during prediction is uncertain, and latency is affected by vendor factors. Therefore, we run a small-sample calibration (2-4 queries per model) to obtain rough estimates of prediction cost and inference latency. When local deployment does not allow such calibration, we use the activated parameters of the model to estimate its latency. Then, we estimate its token-cost by assuming that conventional LLMs produce comparable completion lengths, while LRMs produce longer outputs by a task-dependent factor $\gamma_{\text{task}}>1$ (\ie the expected output length of an LRM is approximated as $\gamma_{\text{task}}$ times that of a conventional LLM on the same task). The value of $\gamma_{\text{task}}$ can be collected from recent research~\citep{benchmarkingLRM}.

Given the resulting triplets $\mathbf{z}_m=\langle \mathrm{Perf}_m,\mathrm{TokCost}_m,\mathrm{Lat}_m\rangle$ for all LLMs after applying the Pareto frontier, we perform $k$-medoids clustering with $\big[\mathrm{Perf}_m,\ \log \mathrm{TokCost}_m,\ \log \mathrm{Lat}_m\big]$. We apply $log$ here because we aim to group LLMs with similar budget levels (\eg  the token-cost and latency gap between LLM in a pool should be kept within $5\times$). After clustering, we downsample overpopulated clusters by removing items farthest from the center, and we supplement undersized clusters with their nearest neighbors. This balancing enables an effective query-conditioned role-backbone matching to allocate resources to agents.

We present the LLM pool for each dataset in Table~\ref{tab:LLM_pools}. MMLU-ind1 and MMLU-ind2 represent the settings of remove LLMs and add new LLMs in inductive analysis respectively (Section~\ref{sec:inductive}).

\textbf{Backbone profiling.}\quad
For each model $m$ among the candidate LLMs, we construct three \emph{profiles}. 
(i) \textit{performance profile}: it summarizes the model’s benchmark performance and reflects the model’s capability for subsequent matching like~\citep{graphrouter}. Benchmark performance is taken from official reports~\citep{qwen3}. 
(ii) \textit{PTP profile}: it contains the per-token price (PTP) information as shown in Table~\ref{tab:LLM-price}. 
(iii) \textit{type profile}: it provides the information on whether the model is a conventional LLM or an LRM. 
The profiles will then be encoded by the $\mathrm{TextEncoder}$ into three LLM embeddings $e_m^{\mathrm{perf}},\,e_m^{\mathrm{ptp}},\,e_m^{\mathrm{type}}$.

\textbf{Templates of LLM Profiles}\quad
\label{appendix:template}
We demonstrate the profiles for Qwen3-32B (non-thinking version). We only include the benchmark results that are related to the current task in the performance profile to reduce redundant information. In the PTP profile, we present the per-token price of the LLM. In the type profile, we present the description of a conventional LLM. We also provide the type profile for LRMs: 
\\
\begin{profilebox}[colframe=RoyalBlue, colback=RoyalBlue!6, colbacktitle=RoyalBlue]{Performance Profile}
The Qwen3-32B multilingual large language model (LLM) is a pretrained and instruction tuned generative model with 32 billion parameters.

Its benchmark result for AgentCoding is : BFCL v3 63.00 ; LiveCodeBench v5 31.30 ; CodeForces 1353.
\end{profilebox}

\begin{profilebox}[colframe=ForestGreen, colback=ForestGreen!6, colbacktitle=ForestGreen]{PTP Profile}
This model has an per-token price of : 1 CNY per million input tokens,  4 CNY per million output tokens.
\end{profilebox}

\begin{profilebox}[colframe=DarkGray, colback=DarkGray!6, colbacktitle=DarkGray]{Type Profile (For Conventional LLM)}
Standard inference mode with balanced performance, latency, and cost. 

No capability of deep reasoning, but fast in response generation and low cost and latency
\end{profilebox}

\begin{profilebox}[colframe=DimGray, colback=DimGray!6, colbacktitle=DimGray]{Type Profile (For LRM)}
Deep thinking mode that enables more deliberate internal reasoning steps during inference.

This mode typically results in longer token usage, significantly increasing both token usage cost and large response latency.
\end{profilebox}

\begin{table*}[t]
\centering
\scriptsize % 整体缩小字体
\setlength{\tabcolsep}{4pt}
\renewcommand{\arraystretch}{1.2}

\caption{LLM pools used by \textsc{AgentBalance} across datasets. Pools are ordered from weaker (Pool 0) to stronger (Pool 3). Qwen3 models with thinking mode (\ie reasoning type) are suffixed with '-thinking'.}
\vspace{-0.3cm}
\label{tab:LLM_pools}
\begin{tabular}{l p{2.7cm} p{2.7cm} p{2.7cm} p{2.7cm}}
\toprule
\textbf{Dataset} & \textbf{Pool 0 (weak)} & \textbf{Pool 1} & \textbf{Pool 2} & \textbf{Pool 3 (strong)} \\
\midrule
HumanEval &
Qwen3-1.7B \newline Qwen3-8B \newline Qwen3-14B &
Qwen3-14B \newline Qwen3-32B \newline Qwen3-235B-A22B &
Qwen3-1.7B-thinking \newline Qwen3-8B-thinking \newline Qwen3-14B-thinking &
Qwen3-14B-thinking \newline Qwen3-32B-thinking \newline Qwen3-235B-A22B-thinking \\
\midrule
MMLU &
Qwen3-1.7B \newline Qwen3-1.7B-thinking \newline Qwen3-8B &
Qwen3-14B \newline Qwen3-8B-thinking \newline Qwen3-14B-thinking &
Qwen3-235B-A22B \newline Qwen3-14B-thinking \newline Qwen3-32B-thinking &
Qwen3-235B-A22B \newline Qwen3-32B-thinking \newline Qwen3-235B-A22B-thinking \\
\midrule
MATH &
Qwen3-1.7B \newline Qwen3-8B \newline Qwen3-1.7B-thinking &
Qwen3-14B \newline Qwen3-1.7B-thinking \newline Qwen3-8B-thinking &
Qwen3-235B-A22B \newline Qwen3-14B-thinking \newline Qwen3-32B-thinking &
Qwen3-235B-A22B \newline Qwen3-32B-thinking \newline Qwen3-235B-A22B-thinking \\
\midrule
MMLU-ind1 &
Qwen3-1.7B \newline Qwen3-1.7B-thinking \newline Qwen3-8B &
Qwen3-1.7B-thinking \newline Qwen3-8B-thinking \newline Qwen3-14B-thinking &
Qwen3-8B \newline Qwen3-14B \newline Qwen3-14B-thinking &
Qwen3-8B-thinking \newline Qwen3-14B-thinking \newline Qwen3-32B-thinking \\
\midrule
MMLU-ind2 &
Qwen3-1.7B \newline Qwen3-1.7B-thinking \newline Qwen3-8B &
Qwen3-1.7B-thinking \newline Qwen3-8B \newline Qwen3-8B-thinking &
Qwen3-14B \newline DeepSeek-V3 \newline Qwen3-Next-80B-A3B &
Qwen3-14B-thinking \newline DeepSeek-R1 \newline Qwen3-235B-A22B-thinking \\
\bottomrule
\end{tabular}
\label{tab:LLM_pools}
\vspace{-0.2cm}
\end{table*}

\begin{table}[t]
\centering
\small
\setlength{\tabcolsep}{6pt}
\caption{Hyperparameters of \textsc{AgentBalance} per dataset}
\vspace{-0.3cm}
\begin{tabular}{l r r r r}
\toprule
Dataset & upper bound & $\lambda_{\mathrm{tok}}$ & $\lambda_{\mathrm{lat}}$ & $\delta$ \\
\midrule
\multirow{4}{*}{\emph{HumanEval}}
 & 1 & 100 & $1\times10^{-2}$ & 0.3 \\
 & 2 & 10  & $5\times10^{-3}$ & 0.3 \\
 & 3 & 1   & $1\times10^{-3}$ & 0.4 \\
 & 4 & 0   & $0$              & 0.6 \\
\midrule
\multirow{4}{*}{\emph{MMLU}}
 & 1 & 40  & $5\times10^{-3}$ & 0.3 \\
 & 2 & 10  & $1\times10^{-3}$ & 0.3 \\
 & 3 & 2   & $5\times10^{-4}$ & 0.4 \\
 & 4 & 0   & $0$              & 0.6 \\
\midrule
\multirow{4}{*}{\emph{MATH}}
 & 1 & 20  & $3\times10^{-3}$ & 0.3 \\
 & 2 & 5   & $1\times10^{-3}$ & 0.3 \\
 & 3 & 1   & $2\times10^{-4}$ & 0.4 \\
 & 4 & 0   & $0$              & 0.6 \\
\bottomrule
\end{tabular}
\vspace{-0.4cm}
\label{tab:hparams_usercap}
\end{table}

\begin{algorithm}[t]
\caption{\textsc{AgentBalance}: High-Level Workflow}
\label{alg:agentbalance}
\SetKwInOut{Input}{Input}
\Input{Dataset $\mathcal{D}$ with queries; base framework $M=(G^\ast,\mathcal{B})$ with role templates $T$ and candidate backbones $\mathcal{B}$;
difficulty estimator $f_{\mathrm{diff}}$; penalties $\lambda_{\mathrm{tok}},\lambda_{\mathrm{lat}},\lambda_{\mathrm{len}}$;
difficulty offset $\delta$.}
\BlankLine
\textbf{Pool Construction with Profiling (Initialization)}\;
$P^\ast \leftarrow \textsc{ConstructPools}(\mathcal{B})$\;
$P \leftarrow \textsc{BackboneProfiling}(\mathcal{B} , P^\ast)$\tcp*[r]{Pareto \& clustering; profiling backbones}
\BlankLine
\ForEach{$Q \in \mathcal{D}$}{
  \textbf{Pool Selection}\;
  $d \leftarrow \textsc{EstimateDifficulty}(f_{\mathrm{diff}}, Q)$\;
  $s^\star,\, p_{\mathrm{sel}} \leftarrow \textsc{SelectPool}(P, d, \delta)$\tcp*[r]{$s^\star$: pool index}
  \BlankLine
  \textbf{Role-Backbone Matching}\;
  $\{m_i\},\, p_{\mathrm{match}} \leftarrow \textsc{RoleBackboneMatch}(T, P_{s^\star}, Q)$\;
  $V_Q \leftarrow \textsc{InstantiateAgents}(T, \{m_i\})$\;
  \BlankLine
  \textbf{Agent Representation Learning}\;
  $H \leftarrow \textsc{LearnAgentRepresentations}(V_Q, Q)$\;
  \BlankLine
  \textbf{Agent Gating}\;
  $G,\, p_{\mathrm{gate}} \leftarrow \textsc{GateAgents}(H)$\tcp*[r]{$G$: agent-wise gating mask}
  $V_Q^{\mathrm{gated}} \leftarrow \textsc{ApplyGates}(V_Q, G)$\tcp*[r]{keep $\ge 2$ agents}
  \BlankLine
  \textbf{Latency-Aware Topology Synthesis}\;
  $L_{\max} \leftarrow \textsc{PredictHopBudget}(H)$\;
  $E,\, p_{\mathrm{topo}} \leftarrow \textsc{SynthesizeTopology}(V_Q^{\mathrm{gated}}, H)$\;
  $E \leftarrow \textsc{PruneToLatency}(E, L_{\max})$\;
  \BlankLine
  \textbf{Execute \& Update}\;
  $y,\, \mathrm{Perf},\, \mathrm{Tok},\, \mathrm{Lat} \leftarrow \textsc{ExecuteMAS}(V_Q^{\mathrm{gated}}, E, Q)$\;
  $D \leftarrow \{s^\star, \{m_i\}, G, E\}$\tcp*[r]{decision tuple}
  $\textsc{Update}(\theta;\, D,\, \mathrm{Perf},\, \mathrm{Tok},\, \mathrm{Lat};\, \lambda_{\mathrm{tok}}, \lambda_{\mathrm{lat}}, \lambda_{\mathrm{len}})$\;
  \tcp*[r]{$p_\theta(D|Q)=p_{\mathrm{sel}}\,p_{\mathrm{match}}\,p_{\mathrm{gate}}\,p_{\mathrm{topo}}$}
}
% \vspace{-0.4cm}
\end{algorithm}

\section{Details of Implementation}
\label{appendix:implementation}
\textbf{AgentBalance.}\quad
To evaluate performance under different resource budgets, we adjust hyperparameters of \textsc{AgentBalance} as summarized in Table~\ref{tab:hparams_usercap}. We fix the length penalty at \(\lambda_{\text{len}}=0.2\) and use the configuration with \texttt{upper bound}\(=2\) unless otherwise specified.\\
\textbf{Baselines.}\quad
For single-LLM baselines (\ie AgentPrune, G-designer and AgentDropout), we instantiate the MAS with a single LLM and, for each candidate LLM in turn, evaluate and record the triplets  \((\text{Perf}, \text{token-cost}, \text{latency})\).
For a chosen budget type (token-cost or latency), we construct the budget–performance {frontier} by taking the upper envelope (convex hull) of the \((\text{budget}, \text{performance})\) point.
This frontier provides the maximal performance attainable by methods at different budgets and is used for our Performance-at-Budget comparison.
For MasRouter, we vary $\lambda$ within $\{0,5,40, 100\}$ to launch four experiments similar to \textsc{AgentBalance}.

\section{Details of Metrics}
\label{appendix:metrics}

\paragraph{Performance-at-Budget.}
We present the formulation of Performance-at-Budget (P@$\cdot$)~\citep{pk} here.
Let \(\mathcal{H}=\{(b_j, p_j)\}_{j=1}^{J}\) denote the frontier points for a given budget type, sorted by budget \(b_j\).
For any target budget \(B\) with \(b_\ell \le B \le b_{\ell+1}\), Performance-at-Budget is obtained by linear interpolation between adjacent frontier points:
\begin{equation}
P(B)\;=\;p_\ell\;+\;\frac{B-b_\ell}{\,b_{\ell+1}-b_\ell\,}\,\bigl(p_{\ell+1}-p_\ell\bigr),
\end{equation}
Budgets in (P@$\cdot$) are applied separately for token-cost \(P_{\mathrm{T}}(B_{\mathrm{tok}})\) and latency \(P_{\mathrm{L}}(B_{\mathrm{lat}})\).

\paragraph{Area Under Curve (AUC).}
We quantify the global cost-effectiveness by integrating the area under each method’s performance—budget frontier.
As all these methods are sampled at different budget points, we first convexify each method’s \((\text{budget},\text{performance})\) scatter and use the upper envelope (piecewise linear) as its frontier.
For comparability, we (i) anchor every frontier at \((0,0)\); (ii) define a \emph{common integration window} \([0, B_{\max}^{\text{base}}]\), where \(B_{\max}^{\text{base}}\) is the base MAS’s maximum budget; and (iii) \emph{ensure the right endpoint at \(B_{\max}^{\text{base}}\) is included in the area}: we append the base MAS’s measured point \(\bigl(B_{\max}^{\text{base}}, p_{\max}^{\text{base}}\bigr)\) to every method’s frontier before integration, to control the influence of the high-cost region.
We then integrate the area under each truncated frontier over \([0, B_{\max}^{\text{base}}]\), reporting AUC separately for token-cost and for latency (higher is better within the shared budget range).

\section{Workflow of \textsc{AgentBalance}}
We present the high-level algorithm workflow in Algorithm~\ref{alg:agentbalance}.

\section{Difficulty Estimator}
\label{appendix:difficultyestimator}
We derive supervision from RouterBench~\citep{routerbench}, which aggregates eight domain-specific datasets where each example includes correctness outcomes from multiple LLMs. We define an \emph{ease} score
$p_i=\tfrac{1}{M}\;\sum_{m=1}^{M}\mathbb{I}\{\text{LLM}_m \text{ answers } x_i \text{ correctly}\}\in[0,1]$ for an example $x_i$ evaluated by $M$ LLMs (a lower $p_i$ indicates higher difficulty). Then, the estimator takes the question $x_i$ as input to learn from these targets.

We rebalance the training split to maintain an approximate ratio of
\({Agent Coding}:{Math Text Reasoning}:{General Tasks}=1:1:4\).
Our difficulty estimator model is an MPNet~\citep{mpnet} encoder followed by a lightweight MLP head.
We fine-tune only the last four transformer layers of MPNet, using a learning rate of \(1\!\times\!10^{-5}\) for the MPNet parameters and \(5\!\times\!10^{-4}\) for the MLP head, with weight decay \(1\!\times\!10^{-3}\).
We predict an ease score $s_i\in[0,1]$ for each example $x_i$ and train against soft targets $p_i$. We minimize the per-example average of two equally weighted terms: binary cross-entropy $\mathrm{BCE}(s_i,p_i)$ (with soft targets) and mean-squared error $\mathrm{MSE}(s_i,p_i)$. The resulting model provides an initial difficulty signal, which is subsequently fine-tuned within our difficulty-aware pool selection module to  align with cost-effectiveness objectives.

\section{Evaluation Budgets}
\label{appendix:budgets}
We place the evaluation budgets for the main experiments in Table~\ref{tab:ab_budgets_main} and budgets for the plug-in experiments in Table~\ref{tab:plugin_budgets}. The token-cost is calculated as the total token expenditure for answering questions in evaluation. The latency here is the recorded average latency for questions in the evaluation.
\begin{table}[t]
\centering
\scriptsize
\setlength{\tabcolsep}{4pt}
\caption{Budgets used for the main experiments (Table~\ref{tab:ab_one_table}). $P@T_k$ are token-cost budgets and $P@L_k$ are latency budgets.}
\begin{tabular}{l r r r r r r r r}
\toprule
& \multicolumn{4}{c}{\textbf{Token-Cost Budgets (USD)}} & \multicolumn{4}{c}{\textbf{Latency Budgets (s)}} \\
\cmidrule(lr){2-5}\cmidrule(lr){6-9}
Dataset & $P@T_1$ & $P@T_2$ & $P@T_3$ & $P@T_4$  & $P@L_1$ & $P@L_2$ & $P@L_3$ & $P@L_4$\\
\midrule
MMLU      & 0.07 & 0.13 & 0.32 & 0.75 & 19.0 & 42.0 & 63.0 & 135.0 \\
HumanEval & 0.03 & 0.04 & 0.12 & 0.60 &  9.0 & 11.0 & 40.0 &  100.0 \\
MATH      & 0.08 & 0.17 & 0.50 & 1.05 & 28.0 & 51.0 & 100.0 & 210.0 \\
\bottomrule
\end{tabular}
\label{tab:ab_budgets_main}

\vspace{0.4cm}
\centering
\scriptsize
\setlength{\tabcolsep}{4pt}
\caption{Budgets used for the plug-in experiments (Table~\ref{tab:plug_in_effect}).
$P@T_k$ are token-cost budgets and $P@L_k$ are latency budgets.}
\begin{tabular}{l r r r r r r r r}
\toprule
& \multicolumn{4}{c}{\textbf{Token-Cost Budgets (USD)}}
& \multicolumn{4}{c}{\textbf{Latency Budgets (s)}} \\
\cmidrule(lr){2-5}\cmidrule(lr){6-9}
Setting & $P@T_1$ & $P@T_2$ & $P@T_3$ & $P@T_4$  & $P@L_1$ & $P@L_2$ & $P@L_3$ & $P@L_4$\\
\midrule
MMLU-L  & 0.07 & 0.11 & 0.38 & 0.50 & 23.0 & 30.0 & 74.0 & 96.0 \\
MMLU-AG & 0.06 & 0.11 & 0.35 & 0.67 & 19.0 & 36.0 & 62.0 & 96.0 \\
\bottomrule
\end{tabular}
\label{tab:plugin_budgets}
% \vspace{-0.2cm}
\end{table}

\end{document}